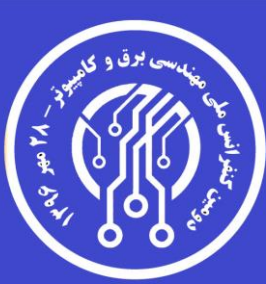
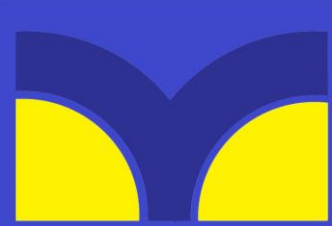

# Content-based image retrieval using Mix histogram

**Mohammad Rezaei[1,*] - Ali Ahmadi[2] - Navid Naderi[3]**
[1] Department of computer engineering, K.N.Toosi University of Technology, Tehran, Iran
*corresponding author: mohmd.rezaei92@email.kntu.ac.ir, telephone number: 00989171704882
[2] Department of computer engineering, K.N.Toosi University of Technology, Tehran, Iran
Email address: ahmadi@eetd.kntu.ac.ir
[3] Department of computer engineering, K.N.Toosi University of Technology, Tehran, Iran
Email address: navid.naderi@email.kntu.ac.ir

**ABSTRACT**

*This paper presents a new method to extract image low-level features, namely mix histogram (MH), for content-based image retrieval. Since color and edge orientation features are important visual information which help the human visual system percept and discriminate different images, this method extracts and integrates color and edge orientation information in order to measure similarity between different images. Traditional color histograms merely focus on the global distribution of color in the image and therefore fail to extract other visual features. The MH is attempting to overcome this problem by extracting edge orientations as well as color feature. The unique characteristic of the MH is that it takes into consideration both color and edge orientation information in an effective manner. Experimental results show that it outperforms many existing methods which were originally developed for image retrieval purposes.*

## 1. INTRODUCTION

By the advent of internet, the amount of digital information began to grow exponentially. As a result, It led to an enormous increase in the volume of digital image data. Image databases have since been and are growing larger and larger. Therefore, there is a growing need for automatic and efficient content-based image retrieval (CBIR) systems. The purpose of these systems is to automatically find similar images to a given image. In a typical CBIR system, in order to measure similarity between images, their feature vectors are compared to each other based on some distance metrics. When a query image comes in, its feature vector will be compared to those in the database and the most similar images will be retrieved. A typical CBIR system diagram can be seen in Figure. 1. CBIR systems often use low-level features (e.g. color, texture and shape) to represent visual features of the image.

Color plays a pivotal role in visual perceptual processes and hence has gained an increasing attention for image retrieval purposes. Color histograms are among common color feature extractors widely used in CBIR systems since they are computationally efficient, easy to implement and invariant to rotation and small changes in viewing position. But it does not take into account the spatial information. Several approaches try to incorporate spatial information such as color coherent vector [1] and color correlograms [2]. Some color feature extractors are also proposed in [3, 4].

Texture is one of the most important visual features of an image widely used in image retrieval applications. There is not yet a clear definition of texture. It is used to characterize roughness or coarseness of object surface. Several approaches have been proposed for texture analysis, such as Markov random field (MRF) [5], Gabor filtering [6, 7], gray co-occurrence matrices [8], the Tamura texture feature [9] and wavelet decomposition [10, 11] and so on.

Shape features are also among important visual features which play a crucial role in recognizing objects and are widely used in content-based image retrieval systems. Some typical shape descriptors include Fourier transforms coefficients, edge curvature and arc length [12, 13].

Some approaches combine multiple features to improve performance [14, 15]. In [16], the texton co-occurrence matrix (TCM) is proposed to use spatial correlation of textons. The multi-texton histogram (MTH) [17] takes the advantages of the co-occurrence matrix and color histograms to utilize both color and texture features. Integrative co-occurrence matrices [12] is another approach that uses color and texture features.

Researchers have recently turned to machine learning techniques for image retrieval applications. One of the most popular methods is to combine CBIR systems with deep learning to take advantage of semantic information of the image as well as visual information [18-20]. In such methods, the outputs of different layers in the Convolutional Neural

Network (CNN) are used to form the final feature vector. Despite the significant performance these methods provide, image retrieval methods using hand-crafted features are still popular because of some advantages they have, such as ease of implementation, interpretability and being far less computationally expensive.

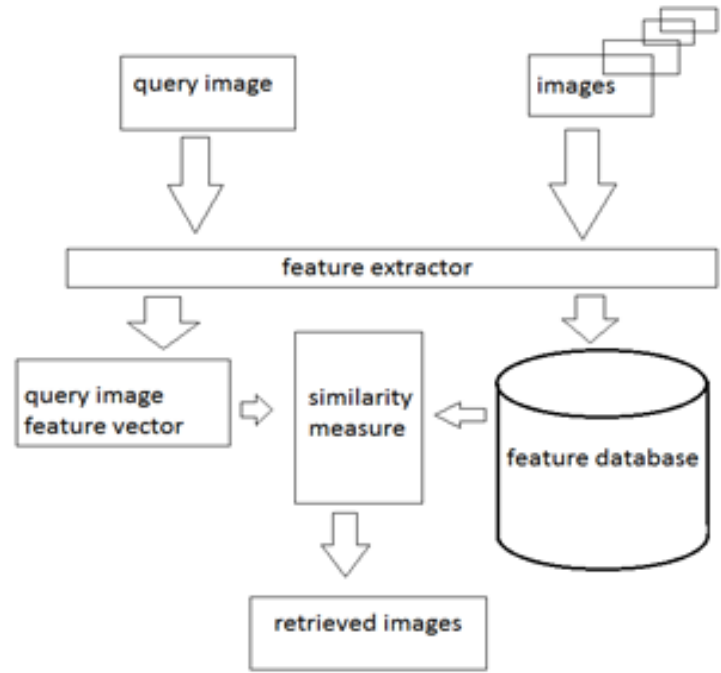

**Fig. 1.** *The diagram of a typical CBIR system*

In this paper, we propose a new method of low-level feature extraction for content-based image retrieval, namely the mix histogram (MH), which captures both color and edge information of images. The MH is a general visual feature descriptor and hence can be used in general purpose image retrieval applications. It is easily implemented and does not include any learning processes or image segmentation.

The remainder of this paper is organized as fallows. In section II, the proposed method is presented. In section III, the performance of the proposed method is evaluated based on comparisons to some benchmarks. Section IV concludes the paper.

**2. THE MIX HISTOGRAM (MH)**

Several studies show that color and edge orientation information play an important role in human image perception [21, 22]. Many approaches have been proposed to extract color or edge orientation features. However, it is still a challenging task to integrate color and edge features in an effective way. In this paper, we propose a method which extracts and integrates color and edge orientation features of the image effectively.
In describing the proposed mix histogram (MH), the HSV color space and color quantization are first briefly described. The detection of edge orientation and orientation quantization in HSV color space is then described, and finally, we introduce the feature representation.

**2.1 HSV Color Space**

A color space is a mathematical model to represent color images. HSV color space is one of the most common color spaces used in digital image processing and machine vision applications. The HSV color space is attempting to mimic the way in which humans perceive color. The HSV stands for the Hue, Saturation, and Value. Hue represents pure color, Saturation is the amount of gray added to the pure color and Value describes the brightness of the color.

The HSV color space is used in the proposed method because the proposed method is a color histogram-based method and in such methods, using the HSV color space results in better retrieval performance when the color space is uniformly quantized. The main reason is that when the color space is uniformly quantized, the number of similar colors which lie in different bins is minimum in the HSV color space compared to other color spaces.

In order to extract color features and facilitate implementation, the color space is first quantized. The task of color quantization is to select a limited set of colors from the whole color space to represent a color image [12]. In this method, we uniformly quantize H channel into 10 bins and S and V channels into 4 bins; therefore $10 \times 4 \times 4 = 160$ color combinations are obtained.

**2.2 Edge Orientation Detection in HSV Color Space**

Edge orientation has a potent influence on early stages of perceptual processes. In this paper, we use the following method for edge orientation detection in the HSV color space.

Color images mostly include 3 color channels, like hue, saturation, and value in the HSV color space. In order to detect edge orientations in an image, the image gradient needs to be computed first. There are many methods to compute color image gradient. One common method is to first convert the color image to a gray-scale image and then compute the image gradient based on the gray-scale image. In this case, much chromatic information will be lost [23]. On the other hand, processing the three individual planes by applying gradient operator separately in each channel to form a composite gradient image can yield erroneous results [13].

In [24], Zenzo proposed an approach for gradient computation of color images. Since the common gradient operators are not applicable to vector functions, the concept of a gradient is extended to the vector maximum rate of change of a scalar function f(x,y) at coordinates (x,y) [13]. In order to compute edge directions in the HSV color space, we use the following algorithm.

Let *h*, *s* and *v* be unit vectors along the H, S and V axis in HSV color space; then, the following vectors are defined for the image [13, 24]:

$$k = \frac{\partial H}{\partial x}h + \frac{\partial S}{\partial x}s + \frac{\partial V}{\partial x}v \qquad (1)$$

$$u = \frac{\partial H}{\partial y}h + \frac{\partial S}{\partial y}s + \frac{\partial V}{\partial y}v \qquad (2)$$

$g_{xx}$, $g_{yy}$ and $g_{xy}$ quantities are defined as follows [13, 24]:

$$g_{xx} = k^T k = \left|\frac{\partial H}{\partial x}\right|^2 + \left|\frac{\partial S}{\partial x}\right|^2 + \left|\frac{\partial V}{\partial x}\right|^2 \qquad (3)$$

$$g_{yy} = u^T u = \left|\frac{\partial H}{\partial y}\right|^2 + \left|\frac{\partial S}{\partial y}\right|^2 + \left|\frac{\partial V}{\partial y}\right|^2 \qquad (4)$$

$$g_{xy} = k^T u = \frac{\partial H}{\partial x}\frac{\partial H}{\partial y} + \frac{\partial S}{\partial x}\frac{\partial S}{\partial y} + \frac{\partial V}{\partial x}\frac{\partial V}{\partial y} \qquad (5)$$

The partial derivatives required for obtaining the above vectors can be computed using any common gradient operator. The sobel operator is used in this paper since it is less sensitive to noise and is computationally good [13].

Let *I(x,y)* be an arbitrary point in the HSV color space. It can be proved that the direction of maximum rate of change of *I(x,y)* is given by the angle [13, 24]:

$$\theta = \frac{1}{2}\tan^{-1}\left[\frac{2g_{xy}}{(g_{xx}-g_{yy})}\right] \qquad (6)$$

and the value of the rate of change at (x,y), in the direction θ, is computed as follows:

$$F(\theta) = \left\{\frac{1}{2}\left[(g_{xx}+g_{yy}) + (g_{xx}-g_{yy})\cos 2\theta + 2g_{xy}\sin 2\theta\right]\right\}^{\frac{1}{2}} \qquad (7)$$

Because $\tan(\alpha) = \tan(\alpha \pm \pi)$, if $\theta_0$ is a solution to (6), so is $\theta_0 \pm \pi/2$. Since (6) gives two different values with a difference of 90º, this equation provides a pair of orthogonal directions for each pixel (x,y). Along one of those directions F is maximum, and it is minimum along the other [13]. The edge orientation for each pixel is the θ which maximizes F in that pixel. We then uniformly quantize the edge orientations into $N_q$ bins. After edge orientation quantization, each pixel can take a value from 1 to $N_q$ as the edge orientation.

### 2.3 Feature Representation

Color and edge orientation are among important visual features which have a direct impact on the human image perception. It is very important to use color feature and take into account edge features of the image at the same time as much as possible. Therefore, in this paper, we represent a new image feature descriptor that uses both color and edge orientation information in an effective way.

Let $I$ be an $n \times n$ image. For a pixel $p = (x,y) \in I$, let $I(p)$ and $G(p)$ denote its color and edge orientation respectively. Let $I_c \triangleq \{p \mid I(p) = c\}$ and $G_q \triangleq \{p \mid G(p) = q\}$. $N_q$ and $N_c$ denote the quantization numbers for edge orientations and colors respectively. The mix histogram (MH) will be a matrix with $N_q$ rows and $N_c$ columns:

$$MH_{i,j} = Pr_{p \in I}\{p \in G_i, I_j\} \qquad (11)$$

In the mix histogram, $MH_{i,j}$ represents the probability with which an arbitrary pixel in image $I$ has the color $j$ and the edge orientation $i$. The MH matrix is eventually converted to a row vector to form the final feature vector. Therefore, the proposed method takes into account both color and edge orientation information in an effective manner. It is clear that if $N_q = 1$, then the mix histogram will be a traditional color histogram.

## 3 EXPERIMENTAL RESULTS

In this section, the Corel-5k dataset is used to evaluate the performance of the proposed method. In the experiments, we randomly selected 20 images from each category as query images, as in [23]. Then the performance was calculated based on the average of results of all queries. In order to have a fair comparison, we selected methods originally designed and implemented for image retrieval, such as color auto-correlograms (CAC) [2] and the multi-texton histogram (MTH) [17]. As discussed in the previous sections, we also set the parameters related to experiments in such a way so as to have an exact and fair comparison.

### 3.1 Datasets

A large number of datasets are used for evaluation of the performance of image retrieval methods such as Corel datasets and Caltech 101 dataset and so on. Among these databases, some are mostly used for testing retrieval performance of methods which are developed for texture analysis while some are mostly used for methods which are developed for color analysis. We selected corel-5k dataset since it contains various images which are rich in both texture and color. It contains 50 categories covering 5000 various images. Each category contains 100 images of size $192 \times 128$ or $128 \times 192$ in JPEG format.

### 3.2 Distance and Performance Metric

Experiments showed that using the distance metric used in [23] yields the best performance retrieval in the proposed method. However, it should be emphasized that it does not mean that this distance metric is always the best choice for all other image retrieval methods. Suppose an image feature vector $T = [T_1, T_2, \ldots, T_M]$ stored in the database needs to be compared to a given query image $Q = [Q_1, Q_2, \ldots, Q_M]$, where M is the size of the feature vectors. Then the distance metric between them is calculated as follows [23]:

$$D(T, Q) = \sum_{i=1}^{M} \frac{|T_i - Q_i|}{|T_i + u_r| + |Q_i - u_Q|} \qquad (12)$$

where $u_r = \sum_{i=1}^{M} T_i/M$ and $u_Q = \sum_{i=1}^{M} Q_i/M$.

In our experiment, we used the Precision and Recall curve, a performance criterion commonly used in information retrieval [25, 26]. We define the Precision and Recall like [23], as follows:

$$P(N) = \frac{I_N}{N} \qquad (13)$$

$$R(N) = \frac{I_N}{M} \qquad (13)$$

Where $I_N$ is the number of relevant images retrieved as $N$ most similar images to the query image, M is the total number of relevant images in the database and $N$ is the total number of images retrieved. In our system, $N = 12$ and $M = 100$. It should be recalled that the value of $N$ affects the final precision value of methods which are being evaluated based on (13). Thus, as mentioned previously, *N* is set 12 as in [23] in order to have a fair comparison. It is can easily be shown that higher precision and recall shows better retrieval performance.

### 3.3 Retrieval Performance

Different quantization numbers for color and edge orientation are used to evaluate the performance of the MH. The values of precision are listed in Table 1. It can be seen that the MH yields the best performance when the quantization number for edge orientation and color are 4 and 160 respectively. Since feature vector size equals $N_q \times N_c$, this results in feature vectors of size $4 \times 160 = 640$. It should be stressed that the greater color or orientation quantization number does not necessarily result in a better performance. For instance, if the color quantization number increases too much, the performance will be reduced because the algorithm will be more affected by noise.

**Table 1.** The performance of the MH with different quantization numbers for edge orientation and color

| The quantization number for edge orientation $N_q$ | The quantization number for color $N_c$ | | | |
|---|---|---|---|---|
| | 72 | 90 | 160 | 240 |
| 3 | 49.99 | 50.73 | 52.06 | 50.69 |
| 4 | 50.97 | 51.72 | 52.80 | 51.38 |
| 5 | 50.90 | 51.40 | 52.24 | 50.87 |

The average precision and recall curve is plotted in Figure. 2. The horizontal axis corresponds to the recall and the vertical axis corresponds to the precision. It can be seen form Figure. 2 and Table 2. that the proposed method outperforms the CAC and MTH algorithms.

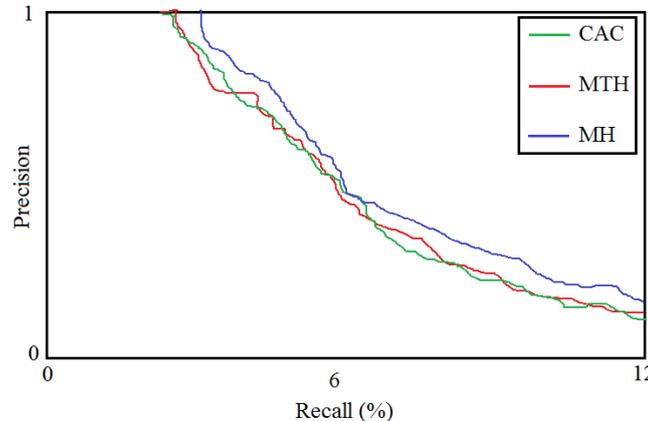

**Fig. 2.** The precision and recall curve of CAC, MTH and MH algorithms

**Table 2.** The average Precision and recall of methods

| Performance | Method | | |
|---|---|---|---|
| | CAC | MTH | MH |
| Precision (%) | 49.06 | 49.84 | 52.80 |
| Recall (%) | 5.9 | 5.98 | 6.33 |

Two retrieval examples can be seen in Figure. 3 and 4. Each figure shows the images considered as the 12 most similar images to the query image (the top-left image is the query image). The numbers above the images denote the category to which they belong.

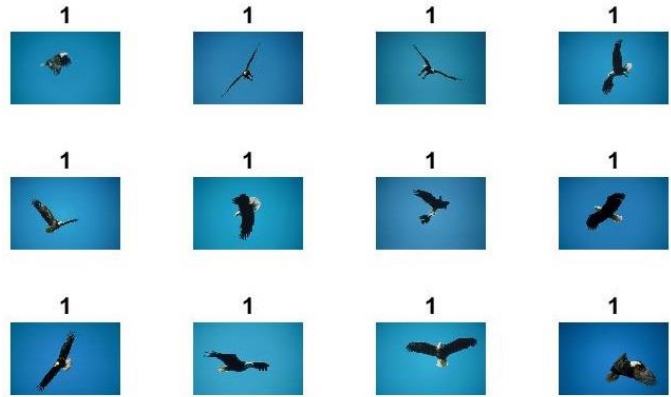

**Fig. 3.**   A retrieval example using the MH

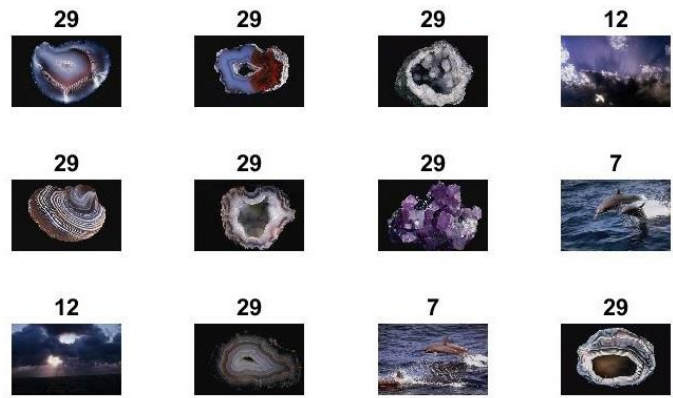

**Fig. 4.**   A retrieval example using the MH

## 4  CONCLUSION

In this paper, we present a new method to extract image low-level features, namely mix histogram (MH), for content-based image retrieval. This method characterizes an image by its color distribution and edge orientations. Traditional color histograms merely focus on the global distribution of color in the image and do not take into account the edge information of the images. Therefore, the mix histogram is different from the traditional color histograms. This feature extraction method does not include any learning process, image segmentation or preprocessing. Our experiments show that the proposed method outperforms most of the image retrieval known algorithms. The mix histogram is powerful but needs to be more investigated in further details.

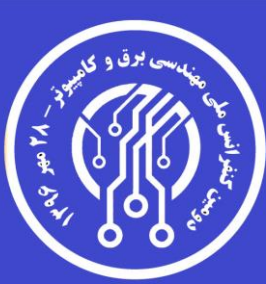
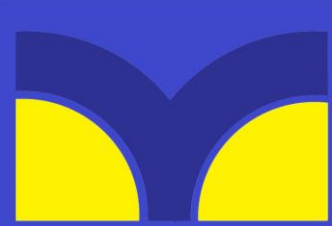